\documentclass[conference]{IEEEtran}
\usepackage{amsmath,amssymb,amsfonts}
\usepackage{graphicx}
\usepackage{listings}
\usepackage{subfigure}
\usepackage[bookmarks=false]{hyperref}

\def\BibTeX{{\rm B\kern-.05em{\sc i\kern-.025em b}\kern-.08em
    T\kern-.1667em\lower.7ex\hbox{E}\kern-.125emX}}
\begin{document}

\title{A Comparative Analysis of Forecasting Financial Time Series Using ARIMA, LSTM, and BiLSTM}

\author{
\IEEEauthorblockN{Sima Siami-Namini} 
\IEEEauthorblockA{Department of Math and Statistics \\
Texas Tech University\\
Email: sima.siami-namini@ttu.edu}
\and
\IEEEauthorblockN{Neda Tavakoli} 
\IEEEauthorblockA{Department of Computer Science \\
Georgia Institute of Technology\\
Email: neda.tavakoli@gatech.edu}
\and
\IEEEauthorblockN{Akbar Siami Namin} 
\IEEEauthorblockA{Department of Computer Science \\
Texas Tech University\\
Email: akbar.namin@ttu.edu}}

\maketitle

\begin{abstract}
Machine and deep learning-based algorithms are the emerging approaches in addressing prediction problems in time series. These techniques have been shown to produce more accurate results than conventional regression-based modeling. It has been reported that artificial Recurrent Neural Networks (RNN) with memory, such as Long Short-Term Memory (LSTM), are superior compared to Autoregressive Integrated Moving Average (ARIMA) with a large margin. 
The LSTM-based models incorporate additional ``{\it gates}'' for the purpose of memorizing longer sequences of input data. The major question is that whether the gates incorporated in the LSTM architecture already offers a good prediction and whether additional training of data would be necessary to further improve the prediction.

Bidirectional LSTMs (BiLSTMs) enable additional training by traversing the input data twice (i.e., 1) left-to-right, and 2) right-to-left). The research question of interest is then whether BiLSTM, with additional training capability, outperforms regular unidirectional LSTM. This paper reports a behavioral analysis and comparison of BiLSTM and LSTM models. The objective is to explore to what extend additional layers of training of data would be beneficial to tune the involved parameters. The results show that additional training of data and thus BiLSTM-based modeling offers better predictions than regular LSTM-based models. More specifically, it was observed that BiLSTM models provide better predictions compared to ARIMA and LSTM models. It was also observed that BiLSTM models reach the equilibrium much slower than LSTM-based models.
\end{abstract}

\section{Introduction}
\label{sec:introduction}
Forecasting is an essential but challenging part of time series data analysis. The type of time series data along with the underlying context are the dominant factors effecting the performance and accuracy of time series data analysis and forecasting techniques employed. Some other domain-dependent factors such as seasonality, economic shocks, unexpected events, and internal changes to the organization which are generating the data also affect the prediction. 

The conventional time series data analysis techniques often utilize 1) linear regressions for model fitting, and then 2) moving average for the prediction purposes. The de facto standard of such techniques is ``{\it Auto-Regressive Integrated Moving Average}'', also known as ARIMA. This linear regression-based approach has been evolved over the years and accordingly many variations of this model have been developed such as SARIMA (or Seasonal ARIMA), and ARIMAX (or ARIMA with Explanatory Variable). These models perform reasonably well for short-term forecasts (i.e., the next lag), but their performance deteriorates severely for long-term predictions. 

Machine learning and more notably deep learning-based approaches are emerging techniques in AI-based data analysis. These learning and AI-based approaches take the data analytical processes into another level, in which the models built are data-driven rather than model-driven. With respect to the underlying application domain, the best learning model can be trained. For instance, a convolution-based neural networks (CNNs) is suitable for problems such as image recognition; whereas, the recurrent neural networks (RNNs) better fit to modeling problems such as time series data and analysis. 

There are several variations of RNN-based models. Most of these RNN-based models differ mainly because of their capabilities in remembering input data. In general, a vanilla form of RNN does not have the capability of remembering the past data. In terms of deep learning terminologies, these models are {\it feed forwarding-based} learning mechanisms. A special type of RNN models is the Long Short-Term Memory (LSTM) networks, through which the relationships between the longer input  and output data are modeled. These RNN-based models, called {\it feedback-based} models, are capable of learning from past data, in which several gates into their network architecture are employed in order to {\it remember} the past data and thus build the prospective model with respect to the past and current data. Hence, the input data are traversed only once (i.e., from left (input) to right (output)). 

It has been reported that the deep learning-based models outperform conventional ARIMA-based models in forecasting time series and in particular for the long term prediction problems \cite{S. Namini-2018}. Even though the performance of LSTM has been shown to be superior to ARIMA, an interesting question is whether its performance can be further improved by incorporating additional layers of training data into the LSTM.

To investigate whether incorporating additional layers of training into the architecture of an LSTM improves its prediction, this paper explores the performance of Bidirectional LSTM (BiLSTM). In an BiLSTM model the given input data is utilized twice for training  (i.e., first from left to right, and then from right to left). In particular, we would like to perform a behavioral analysis comparing these two architectures when training their models. To do so, this paper reports the results of an experiment in which the performance and behavior of these two RNN-based architectures are compared. In particular, we are interested in addressing the following research questions:
\begin{enumerate}
    \item Is the prediction improved when the time series data are learned from both directions (i.e., past-to-future and future-to-past)?
    \item How different these two architectures  (LSTM and BiLSTM) treat input data?
    \item How fast these two architectures reach the equilibrium?
\end{enumerate}

To address these questions, this paper conducts a series of experiments and reports the results. In particular, this paper makes the following key contributions:
\begin{itemize}
    \item[--] Investigate whether additional layers of training improve prediction in the financial time series context. 
    \item[--] Provide a performance analysis comparing the prediction's accuracy of the uni-LSTM and its extension, BiLSTM.  The analysis shows that BiLSTM models outperform LSTMs by $37.78\%$ reduction in error rates. 
    \item[--] Conduct a behavioral analysis of learning processes involved in training the LSTM and BiLSTM-based models. According to the results, BiLSTMs train their models differently than LSTMs by fetching smaller batches of data for training. It was also observed that, the BiLSTM models reach the equilibrium slower than uni-LSTMs. 
\end{itemize}

This paper is structured as follows: Section \ref{sec:related} reviews the related works. The essential background and mathematical formulations are given in Section \ref{sec:background}. The procedure for experimental setup, data collection, and preparation is presented in Section  \ref{sec:comparison}. Section \ref{sec:algorithms} presents the pseudo-code of the developed algorithms. The results of the experiments are reported through Section \ref{sec:results}. Section \ref{sec:discussion} discusses the performance of the algorithms while factors are controlled. The conclusion of the paper and the possible future research directions are provided in Section \ref{sec:conclusion}.

\section{Related Works}
\label{sec:related}

Traditional approaches to time series analysis and forecasting are primarily based on Autoregressive Integrated Moving Average (ARIMA) and its many variations such as Seasonal ARIMA (SARIMA) and ARIMA with Explanatory variables (ARIMAX) \cite{Box-1970}. These techniques have been used for a long time in modeling time series problems \cite{Khashei-2011, Alonso-2012, Adebiyi-2014}. While these moving averaged-based approaches perform reasonably well, they also suffer from some limitations \cite{Earnest-2005}:
\begin{itemize}
    \item[--] Since these models are regression-based approaches to the problem, they are hardly able to model data with nonlinear relationships between parameters. 
    \item[--] There are some assumptions about data when conducting statistical tests that need to be held in order to have a meaningful model (e.g., constant standard deviation). 
    \item[--] They are less accurate for long-term predictions.
\end{itemize}


Machine and deep learning-based approaches have introduced new avenue to analyze time series data. 
Krauss et al. \cite{Krauss-2016} used various forms of forecasting models such as deep learning, gradient-boosted trees, and random forests to model S\&P 500 constitutes. 
Krauss et al. also reported that training neural networks and consequently deep learning-based algorithms was very difficult. Lee and Yoo \cite{Lee-2017} introduced an RNN-based approach to predict stock returns. The idea was to build portfolios by adjusting the threshold levels of returns by internal layers of the RNN built. A similar work is performed by Fischera et al.\ \cite{Fischera-2017} for financial data prediction. 

The most similar papers in which the performance of LSTM and its bi-directional variation is compared are \cite{Kim2019, Cui2018}. Kim and Moon report that Bi-directional Long Short-Term Memory model based on multivariate time-series data outperforms uni-directional LSTM. Cui et al.\ \cite{Cui2018} proposed stacking bidirectional and unidirectional LSTM networks for predicting network-wide traffic speed. They report that the stacked architecture outperforms both BiLSTM and uni-LSTMs.

This article is based on the authors previous research work where the performance of ARIMA-based models with the LSTM-based models was compared in the context of predicting economics and financial time series and parameter tuning \cite{S. Namini-2018}, \cite{Tavakoli-2019}. The paper takes an additional step in comparing the performance of three time series modeling standards: ARIMA, LSTM, and BiLSTM. While traditional prediction problems (such as building a scheduler \cite{Tavakoli-IO-2019} and predicting vulnerabilities in software systems \cite{Pang2015}) can benefit largely from bi-directional training, it is unclear whether learning time series data, and in particular financial and economic data, from both sides is beneficial for the purpose of learning. This paper explores this research problem.


\section{Background}
\label{sec:background}

\subsection{Recurrent Neural Networks}\label{RNN}
The Recurrent Neural Networks (RNNs) are an extension of the conventional Feed-Forward neural networks with the ability of managing variable-length sequence inputs. Unlike the conventional Feed-Forward neural networks, which are not generally able to handle sequential inputs and all their inputs (and outputs) must be independent of each others, the RNNs models provide some gates to store the previous inputs and leverage sequential information of the previous inputs. This special RNNs memory is called {\it recurrent hidden states} and gives the RNNs the ability to predict what input is coming next in the sequence of input data. In theory, RNNs are able to leverage previous sequential information for arbitrary long sequences. In practice, however, due to RNNs' memory limitations, the length of the sequential information is limited to only a few steps back. To give a formal definition of RNNs, lets assume $x= (x_1, x_2, x_3, ....,x_T)$ represents a sequence of length $T$, and $h_t$ represents RNN memory at time step $t$, an RNN model updates its memory information using:
\begin{equation} \label{eq1}
h_t = \sigma( W_x x_t + W_h h_{t-1} +b_t)
\end{equation}
where $\sigma$ is a nonlinear function (e.g., logistic sigmoid, a hyperbolic tangent function, or rectified linear unit (ReLU)), $W_x$ and $W_h$ are weight matrices that are used in deep learning model, and $b_t$ is a constant bias.

In general, RNNs have multiple types: one input to many outputs, many inputs to many outputs, and many inputs to one output. In this work, we only consider RNNs that produce one output $y = (y_1, y_2, ...., y_T)$ which is the probability of the next element of a sequence while its previous inputs are given. The sequence probability can be decomposed as following:
\begin{equation}\label{eq2}
\begin{split}
        p(x_1,\dots,x_T)& = p(x_1)  p(x_2 |x_1) p(x_3|x_1, x_2)...\\
        &~~~~p(x_T | x_1, ..., x_{T-1})
\end{split}
\end{equation}
in which each conditional probability distribution is modeled:
\begin{equation}\label{eq3}
p(x_t | x_1, \dots x_{t-1}) = \sigma (h_t)
\end{equation}
 where $h_t$ is calculated using Equation~\ref{eq1}. 

One of the common problems of RNNs is called "{\it vanishing gradients}" which happens when the information about the input or gradient passes thorough a lot of layers, it will vanish and wash out by the time when it reaches to the end or beginning layer. This problem makes it hard for RNNs to capture the long-term dependencies, and as such the training of RNNs will be extremely challenging. Another problem of RNNs, which rarely happens, is called ``{\it exploding gradients},'' which refers to the cases in which information about the input or gradient passes thorough a lot of layers, it will accumulate and result in a very large gradient when it reaches to the end or beginning layer. This problem  makes RNNs hard to train.

 Gradient, which is mathematically defined as partial derivative of output of a function with respect to its inputs, essentially measures how much the output of a function changes with respect to the changes occurred to its inputs. In the  "{\it vanishing gradients}" problem, the RNN training algorithm assigns smaller values to the weight matrix (i.e., a matrix that is used in the process of RNN training) and thus the RNN model stops learning. 
 On the other hand,  in the {\it exploding gradients} problem , the training algorithm assigns higher values to the weight matrix without any reasons. This problem can be solved by truncating/squashing the gradients~\cite{Hochreiter-1997}.

\subsection{Long Short-Term Memory (LSTM) Models}\label{lstm}
As mentioned earlier, RNNs have difficulties in learning long-term dependencies. The LSTM-based models are an extension for RNNs, which are able to address the vanishing gradient problem in a very clean way. The LSTM models essentially extend the RNNs' memory to enable them  keep and learn long-term dependencies of inputs. This memory extension has the ability of remembering information over a longer period of time and thus enables reading, writing, and deleting information from their memories. The LSTM memory is called a ``{\it gated}'' cell, where the word gate is inspired by the ability to make the decision of preserving or ignoring the memory information. An LSTM model captures important features from inputs and preserves this information over a long period of time. The decision of deleting or preserving the information is made based on the weight values assigned to the information during the training process. Hence, an LSTM model learns what information worth to preserve or remove. 

In general, an LSTM model consists of three gates: forget, input, and output gates. The forget gate makes the decision of preserving/removing the existing information, the input gate specifies the extent to which the new information will be added into the memory, and the output gate controls whether the existing value in the cell contributes to the output. 

{\it I) Forget Gate.} A sigmoid function is usually used for this gate to make the decision of what information needs to be removed from the LSTM memory. This decision is essentially made based on the value of $h_{t-1}$ and $x_t$. The output of this gate is $f_t$, a value between $0$ and $1$, where $0$ indicates completely get rid of the learned value, and $1$ implies preserving the whole value. This output is computed as:
    \begin{equation}\label{eq4}
    f_t = \sigma (W_{f_h}[h_{t-1}], W_{f_x}[x_t], b_f)
    \end{equation}
    where $b_f$ is a constant and is called the {\it bias} value.

{\it II) Input Gate.} This gate makes the decision of whether or not the new information will be added into the LSTM memory. This gate consists of two layers: 1) a {\it sigmoid} layer, and 2) a ``$\tanh$'' layer. The sigmoid layer decides which values needs to be updated, and the $\tanh$ layer creates a vector of new candidate values that will be added into the LSTM memory. The outputs of these two layers are computed through: 
    \begin{align}
    i_t     &= \sigma (W_{i_h}[h_{t-1}], W_{i_x}[x_t], b_i) \label{eq5}\\
    c_t^{\texttt{\char`\~}}   &= \tanh(W_{c_h}[h_{t-1}], W_{c_x}[x_t], b_c)\label{eq6}
    \end{align}
    in which $i_t$ represents whether the value needs to be updated or not, and $c_t^{\texttt{\char`\~}}$ indicates a vector of new candidate values that will be added into the LSTM memory. The combination of these two layers provides an update for the LSTM memory in which the current value is forgotten using the forget gate layer through multiplication of the old value (i.e., $c_{t-1}$) followed by adding the new candidate value $i_t * c_t^{\texttt{\char`\~}}$. The following equation represents its mathematical equation:
    \begin{align}
    c_t = f_t * c_{t-1} + i_t * c_t^{\texttt{\char`\~}} \label{eq7}
    \end{align}
    where $f_t$ is the results of the forget gate, which is a value between $0$ and $1$ where $0$ indicates completely get rid of the value; whereas, $1$ implies completely preserve the value.
    

    %
{\it III) Output Gate.} This gate first uses a sigmoid layer to make the decision of what part of the LSTM memory contributes to the output. Then, it performs a non-linear $\tanh$ function to map the values between $-1$ and $1$. Finally, the result is multiplied by the output of a sigmoid layer. The following equation represents the formulas to compute the output:
    \begin{align}
    o_t  &= \sigma (W_{o_h}[h_{t-1}], W_{o_x}[x_t], b_o) \label{eq8}\\ 
    h_t  &= o_t * \tanh (c_t) \label{eq9}
    \end{align}
    where $o_t$ is the output value, and $h_t$ is its representation as a value between $-1$ and $1$.

\subsection{Deep Bidirectional LSTMs (BiLSTM)} 
The deep-bidirectional LSTMs~\cite{Schuster-1997} are an extension of the described LSTM models in which two LSTMs are applied to the input data. In the first round, an LSTM is applied on the input sequence (i.e., forward layer). In the second round, the reverse form of the input sequence is fed into the LSTM model (i.e., backward layer). Applying the LSTM twice leads to improve learning long-term dependencies and thus consequently will improve the accuracy of the model~\cite{baldi1999exploiting}.

\section{LSTM vs. BiLSTM: An Experimental Study}
\label{sec:comparison}

This paper compares the performance of ARIMA, LSTM, and BiLSTM in the context of predicting financial time series. 
\subsection{Data Set}
The authors partially reused the previously collected data \cite{S. Namini-2018}, in which daily, weekly, and monthly  time series of some stock data for the period of Jan 1985 to Aug 2018 were extracted from the Yahoo finance Website\footnote{https://finance.yahoo.com}. The data included 1) Nikkei 225 index (N225), 2) NASDAQ composite index (IXIC), 3) Hang Seng Index (HSI), 4) S\&P 500 commodity price index (GSPC), 5) Dow Jones industrial average index (DJ), and 6) IBM Stock data. The daily IBM stock data were collected for the period of July 2009 to July 2019.


\subsection{Training and Test Data}

The ``Adjusted Close'' variable was chosen as the only feature of financial time series to be fed into the ARIMA, LSTMs and its variation, BiLSTM models. The data set was divided into training and test where 70\% of each data set were used for training and 30\% of each data set was used for testing the accuracy of models. Table \ref{tab:data} provides the statistics of the number of time series' observations.

\begin{table}
\begin{center}
\caption{The time series data studied.}
\label{tab:data}
\setlength{\tabcolsep}{3pt}
\begin{tabular}{|l|r|r|r|}
\hline
\multicolumn{1}{|c|}{\bf Stock} & \multicolumn{2}{|c|}{\bf Observations} & \multicolumn{1}{|c|}{\bf Total} \\ 
\cline{2-3}
&  \multicolumn{1}{|c|}{\bf Train 70\%} & \multicolumn{1}{|c|}{\bf Test 30\%} & \\
\hline
N225.monthly & 283	& 120	& 403 	\\
IXIC.daily & 8,216 & 3,521 & 11,737 \\
IXIC.weekly  & 1,700  & 729 & 2,429 \\
IXIC.monthly  & 390 & 168 & 558 \\
HSI.monthly  & 258 & 110 & 368 \\
GSPC.daily & 11,910 & 5,105 & 17,015 \\
GSPC.monthly & 568 & 243 & 811 \\
DJI.daily & 57,543 & 24,662 & 82,205 \\
DJI.weekly 	& 1,189	& 509	& 1,698	\\
DJI.monthly & 274	& 117	& 391	\\
IBM.daily & 1,762 & 755 & 2,517 \\
\hline
\multicolumn{1}{|c|}{\it Total} & 84,093 & 36,039 & 120,132 \\
\hline
\end{tabular}
\end{center}
\vspace*{-0.3in}
\end{table}

\subsection{Assessment Metrics}
The ``{\it loss}'' values are typically reported by deep learning algorithms. Loss technically is a kind of penalty for a poor prediction. More specifically, the loss value will be zero, if the model's prediction is perfect. Hence, the goal is to minimize the loss values through obtaining a set of weights and biases that minimizes the loss. 
In addition to loss, which is utilized by the deep learning algorithms, researchers often utilize the Root-Mean-Square-Error (RMSE) to assess the prediction performances. RMSE measures the differences between actual and predicated values. The formula for computing RMSE is: 

\begin{equation}
RMSE = \sqrt{\frac{1}{N} \sum_{i=1}^N (y_i -\hat{y}_i)^2}
\end{equation}

Where $N$ is the total number of observations, $y_i$ is the actual value; whereas, $\hat{y}_i$ is the predicated value. The main benefit of using RMSE is that it penalizes large errors. It also scales the scores in the same units as the forecast values. Furthermore, we also used the percentage of reduction in RMSE, as a measure to assess the improvement that can be calculated as:

\begin{equation}
Changes\% = \frac{New\ Value - Original\ Value}{Original\ Value}*100
\end{equation}

\section{The Algorithms}
\label{sec:algorithms}

\begin{figure}
    \centering
    \subfigure[Feedforward]
    {\includegraphics[scale=0.28]{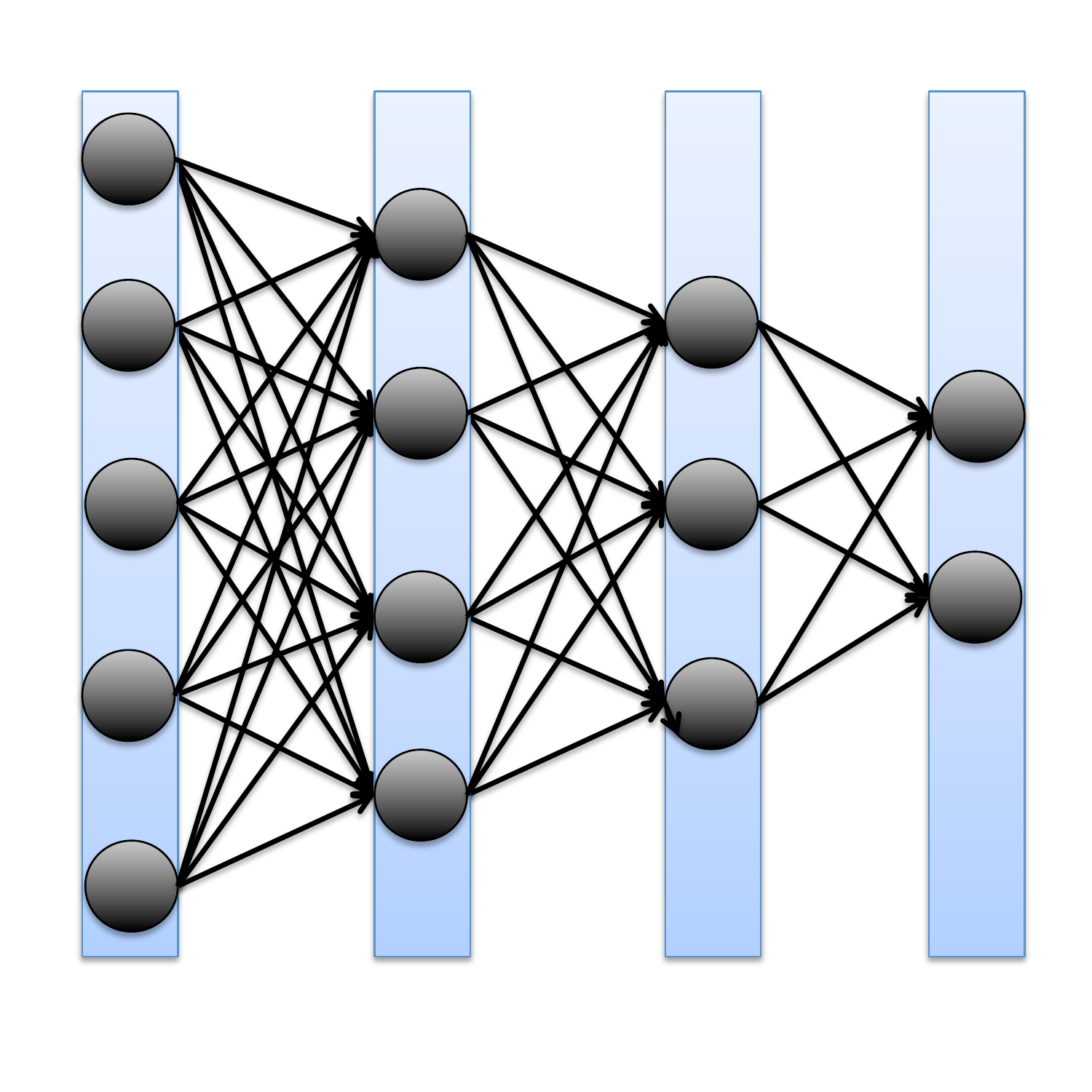}}
    \subfigure[Feedbackward]
    {\includegraphics[scale=0.28]{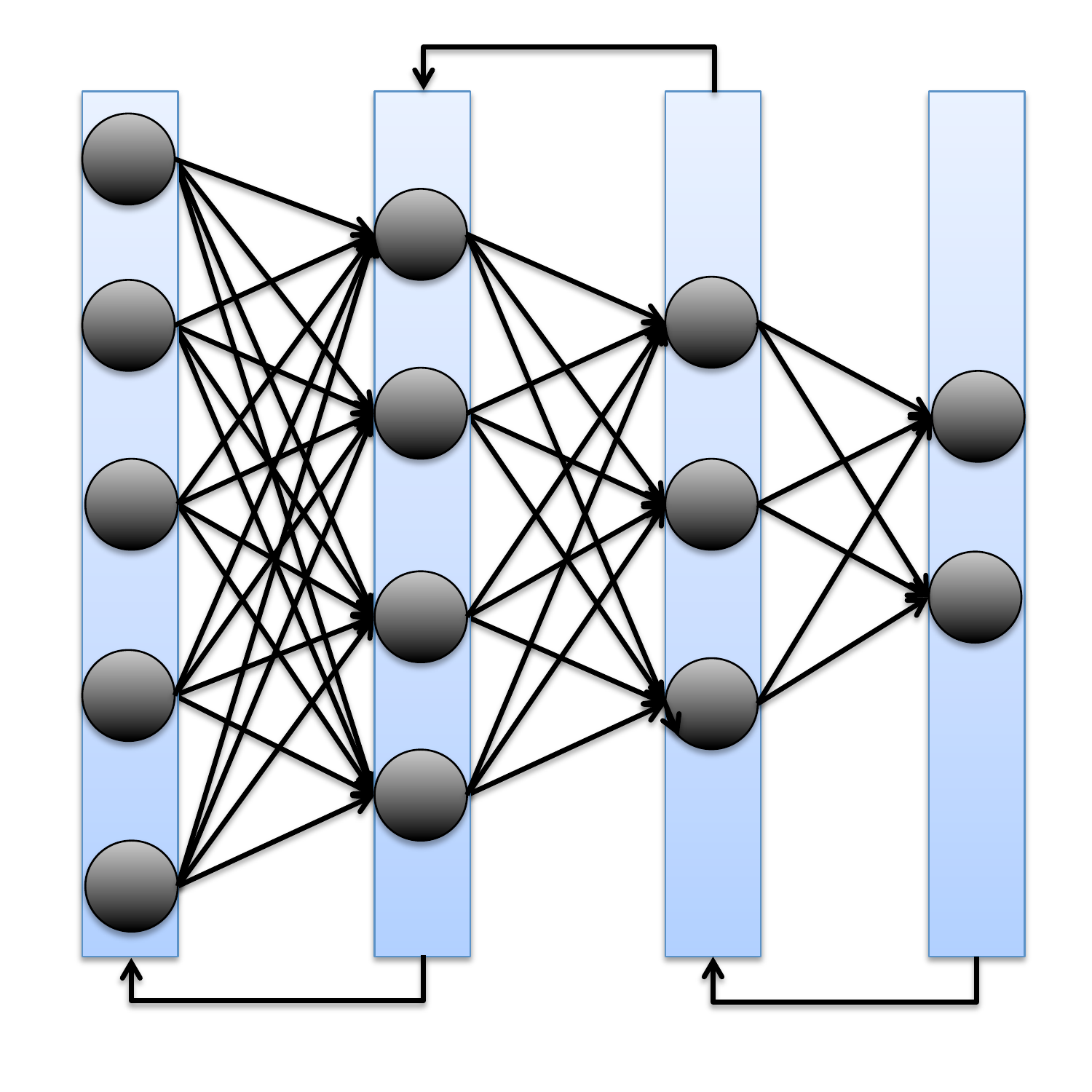}}\\
    \subfigure[LSTM]
    {\includegraphics[scale=0.28]{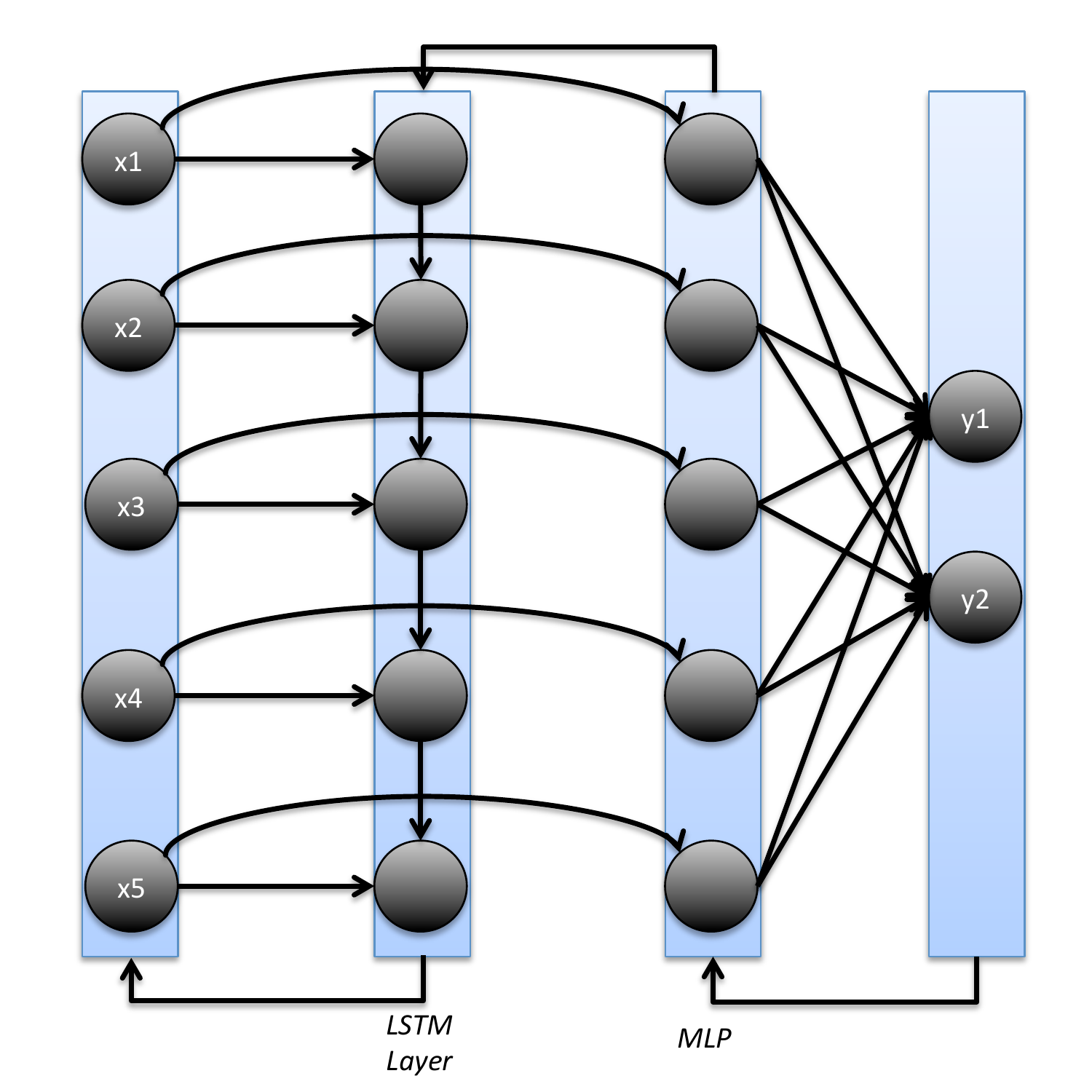}}
    \subfigure[BiLSTM]
    {\includegraphics[scale=0.28]{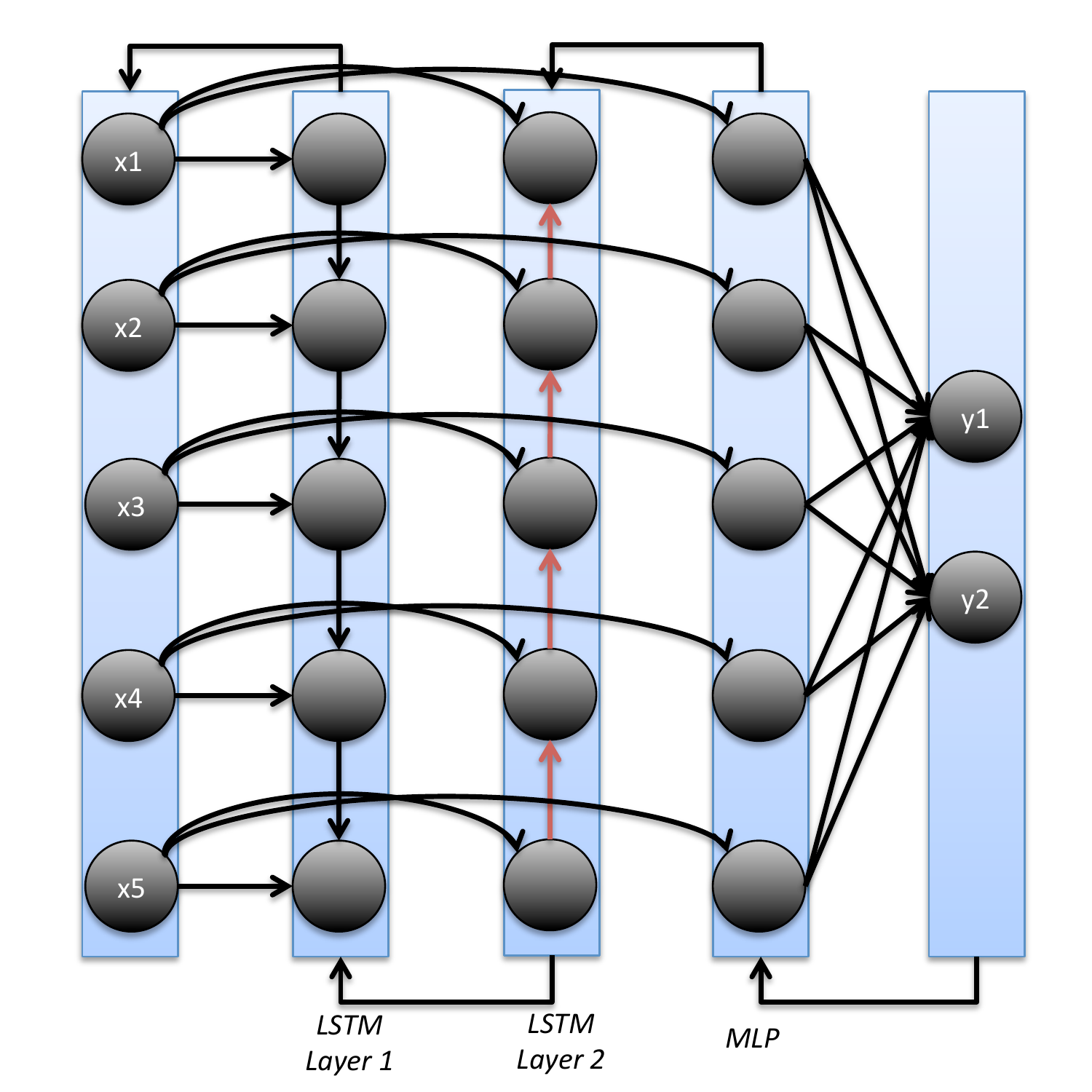}}
    \caption{Various forms of artificial neural networks.}
    \label{fig:architectures}
      \vspace*{-0.25in}
    \end{figure}

The general ``{\it feed-forward}'' Artificial Neural Networks (ANN) (Figure \ref{fig:architectures}(a)) allow training the model by traveling in one direction only without considering any feedback from the past input data. More specifically, ANN models travel directly from input (left) to output (right) without taking into account any feedback from the already trained data from the past. As a result, the output of any layer does not affect the training process performed on the same layer (i.e., no memory). These types of neural networks are useful for modeling the (linear or non-linear) relationship between the input and output variables and thus functionally perform like a regression-based modeling. In other words, through these networks a functional mapping is performed through which the input data are mapped to output data. This type of neural networks is heavily utilized in pattern recognition. The Convolutional Neural Networks (CNN), and the conventional and basic auto-encoder networks are typical ANN models.

On the other hand, the Recurrent-based Neural Networks (RNNs) remember parts of the past data through a methodology, called feedback, in which the training takes place not only from input to output (as feed-forward), but also it utilizes a loop in the network to preserve some information and thus functions like a memory (Figure \ref{fig:architectures}(b)). Unlike feed-forward ANN networks, the feedback-based neural networks are dynamics and their states change continuously until they reach the equilibrium status and are thus optimized. The states remain at the equilibrium status until new inputs are arrived demanding changes in the equilibrium. The major problem of a vanilla form of RNNs is that these types of neural networks cannot preserve and thus does not remember long inputs.  

As an extension to RNNs, Long Short-Term Memory (LSTM) (Figure \ref{fig:architectures}(c)) is introduced to remember long input data and thus the relationship between the long input data and output is described in accordance with an additional dimension (e.g., time or spatial location). An LSTM network remembers long sequence of data through the utilization of several gates such as: 1) input gate, 2) forget gate, and 3) output gate.

The deep-bidirectional LSTMs (BiLSTM) networks ~\cite{Schuster-1997} are a variation of normal LSTMs (Figure \ref{fig:architectures}(d)), in which the desired model is trained not only from inputs to outputs, but also from outputs to inputs. More precisely, given the input sequence of data, a BiLSTM model first feed input data to an LSTM model (feedback layer), and then repeat the training via another LSTM model but on the reverse order of the sequence of the input data (i.e., Watson-Crick complement~\cite{gao2004expanded}). It has been reported that using BiLSTM models outperforms regular LSTMs~\cite{baldi1999exploiting}. The algorithm(s) developed for the experiments reported in this paper are listed in Listing 1. Please note that the two algorithms (LSTM and BiLSTM) are incorporated into one, where lines 9 - 12 switch between the two algorithms. The rolling-based algorithms re-train the models each time a new observation is fetched (line 26). Hence, once a prediction is performed and its value is compared with the actual value, the value is added to the training set (line 26), and the model is re-trained (line 27).

\section{Results}
\label{sec:results}

Table \ref{tab:RMSE} reports the Rooted Mean Squared Error (RMSE) achieved by each technique for forecasting the stock data. In most cases (except IXIC.weekly), a significant reduction in the magnitude of the RMSE values is observed. 

In comparing the LSTM and BiLSTM models, the percentage of reductions varies from $(-)\%77.60$ for DJI.daily to $(-)\%12.93$ for IXIC.daily. On average, the RMSE values achieved for LSTM and BiLSTM-based models are $39.09$ and $20.17$, respectively, and thus achieving $(-)\%37.78$ reduction on average. With respect to the data, it is apparent that BiLSTM models outperform regular uni-LSTM models significantly, with a large margin. 

Table \ref{tab:RMSE} also reports some other results computed for ARIMA and the percentages of the reductions captured. More specifically, the average reductions obtained using BiLSTM over ARIMA is $-93.11$; whereas, the average percentage of reduction using LSTM over ARIMA is reported as $-88.07$. The results indicate that modeling using BiLSTM instead of LSTM and ARIMA indeed improves the prediction accuracy.

\lstset{basicstyle=\small\ttfamily}
\def\listingsfont{\ttfamily} 
\begin{lstlisting}[language=Java, label=verb1,caption={The developed rolling LSTM/BiLSTM algorithms.}, frame=tb, label={prg:box2}, frame=shadowbox, escapeinside=`']
# Rolling LSTM and BiLSTM
Inputs: Time series 
Outputs: RMSE of the forecasted data
# Split data into:
# 70\% training and 30\% testing data
1. size `$\leftarrow$' length(series) * 0.70
2. train `$\leftarrow$' series`['0...size`]'
3. test `$\leftarrow$' series`['size...length(size)`]'
# Set the random seed to a fixed value 
4. set random.seed(7)
5. set option = "L" (LSTM) or "B" (BiLSTM)

# Fit an LSTM or BiLSTM model to training data
Procedure fit_lstm_Bilstm(train, epoch, 
                   neurons, option)
6. X `$\leftarrow$' train
7. y `$\leftarrow$' train - X
8. model = Sequential()
9. if option = 'L':
10.  model.add(LSTM(neurons), stateful=True))
11. else option = 'B':  
12.  model.add(Bidirectional(LSTM(neurons, 
                             stateful=True))

13. model.compile(loss='mean_squared_error', 
                          optimizer='adam')
14.for each i in range(epoch) do
15.  model.fit(X, y, epochs=1, shuffle=False)
16.  model.reset_states()
17.end for 
return model

# Make a one-step forecast
Procedure forecast_lstm(model, X)
18. yhat `$\leftarrow$' model.predict(X)
return yhat

19. epoch `$\leftarrow$' 1
20. neurons `$\leftarrow$' 4
21. predictions `$\leftarrow$' empty
22. lstm_model = fit_lstm_Bilstm(train,epoch,
                             neurons, option) 
# Forecast the training dataset 
23. lstm_model.predict(train)

# Walk-forward validation on the test data
24. for each i in range(length(test)) do
25.    # make one-step forecast
26.    X `$\leftarrow$' test[i]
27.    yhat `$\leftarrow$' forecast_lstm(lstm_model, X)
28.    # record forecast
29.    predictions.append(yhat)
30.    expected `$\leftarrow$' test[i]
31. end for  

32. MSE `$\leftarrow$' mean_squared_error(expected, 
                                predictions)
33. Return (RMSE `$\leftarrow$' sqrt(MSE))
\end{lstlisting}

\begin{figure}
    \centering
    \subfigure[IBM (ARIMA): test data.]
    {\includegraphics[scale=0.4]{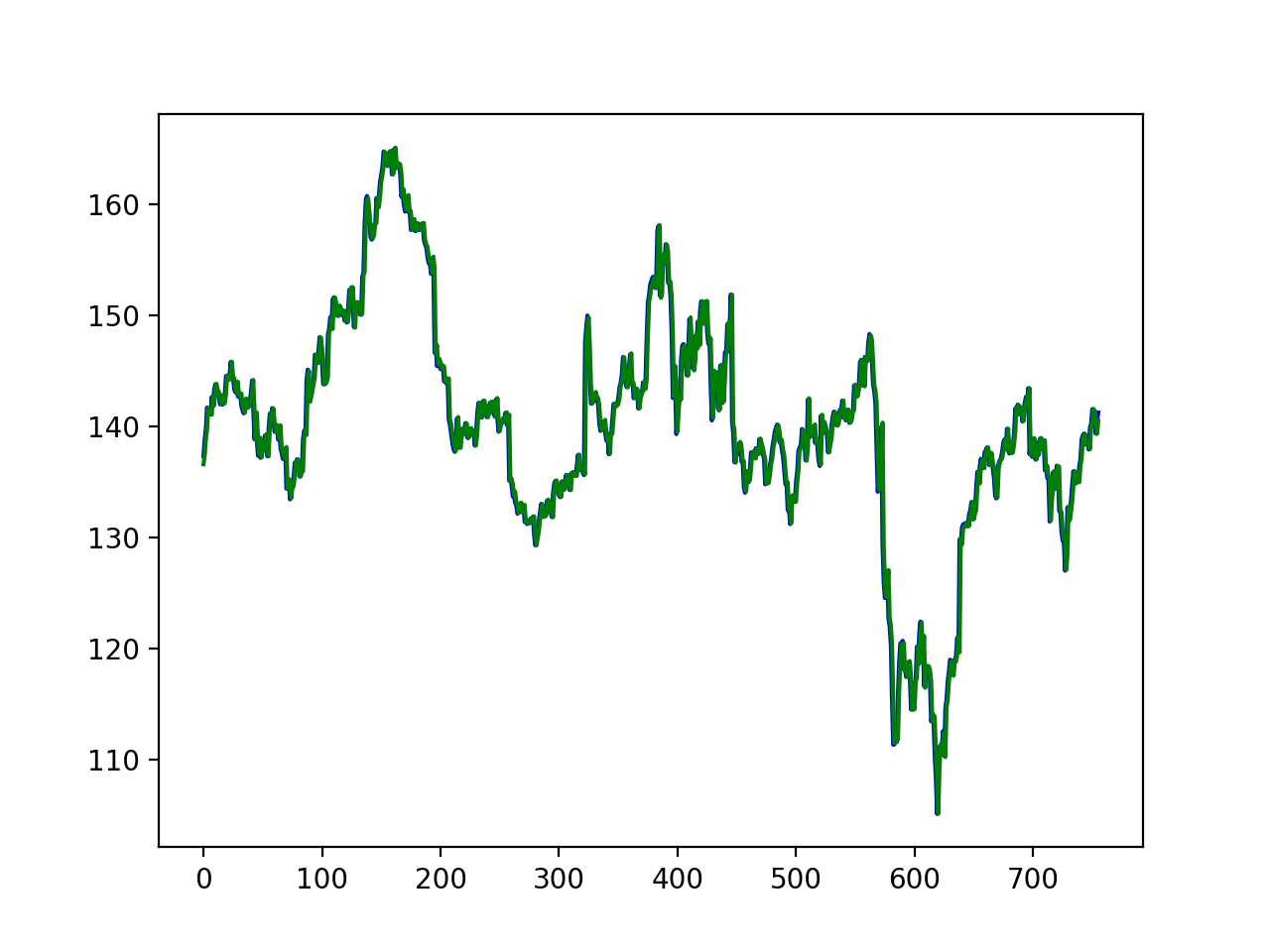}}
    \subfigure[IBM (LSTM): train and test data.]
    {\includegraphics[scale=0.4]{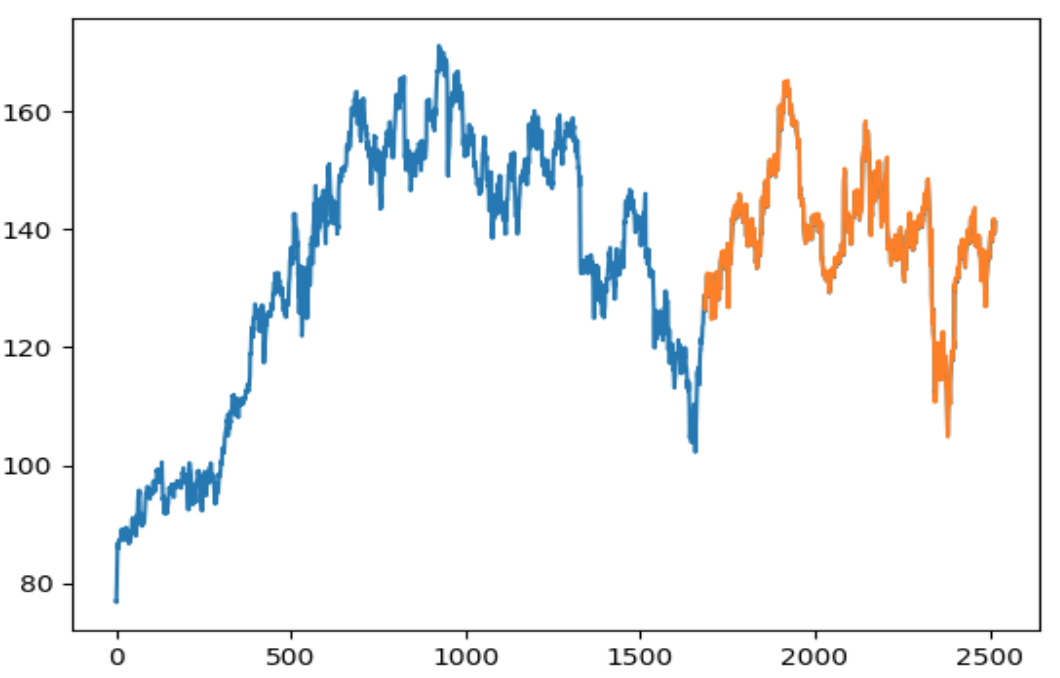}}
    \subfigure[IBM (BiLSTM): train and test data.]
    {\includegraphics[scale=0.4]{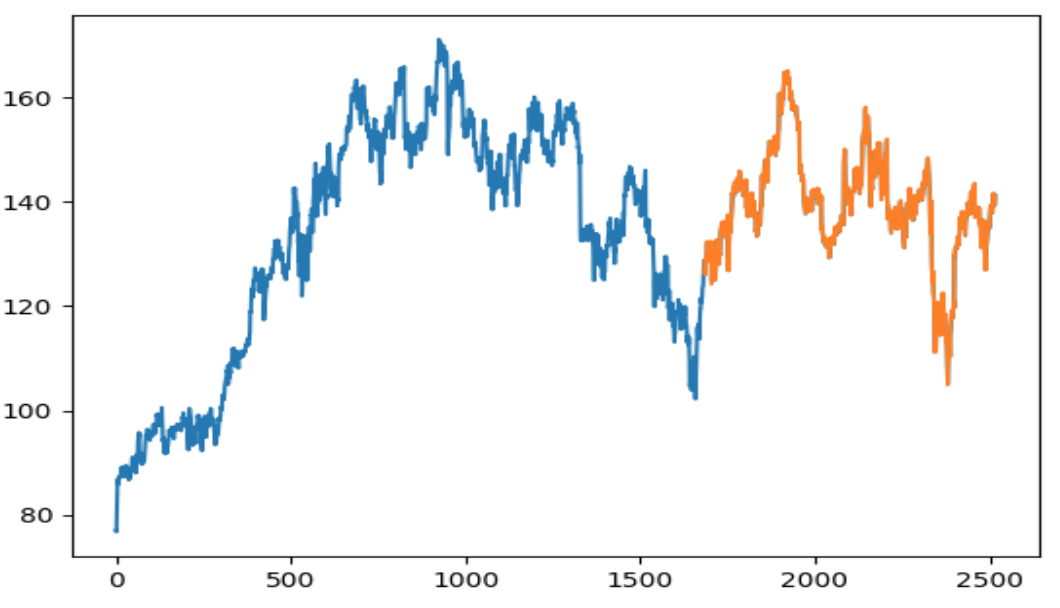}}\\
    \caption{IBM forecasting using ARIMA, LSTM, and BiLSTM.}
    \label{fig:IBM}
       \vspace*{-0.25in}
    \end{figure}

To illustrate the forecasts performed by both LSTM and BiLSTM models,  Figures \ref{fig:IBM}(a)-(c) show the forecasts for the IBM stock estimated by ARIMA, LSTM and BiLSTM, respectively. Please note that the parts colored in green and orange (i.e., predicted parts) are overlapping the original values of test data. As a result, the initial test data are less visible in the plots.

\begin{table}
\begin{center}
\caption{RMSEs of ARIMA, LSTM, and BiLSTM models.}
\label{tab:RMSE}
\setlength{\tabcolsep}{3pt}
\begin{tabular}{|l|r|r|r|r|r|r|}
\hline
 & \multicolumn{3}{|c|}{\bf RMSE} & \multicolumn{3}{|c|}{\bf \% Reduction} \\ 
\cline{2-7}
& \multicolumn{1}{|c|}{\bf } & 
\multicolumn{1}{|c|}{\bf } & \multicolumn{1}{|c|}{\bf Bi} & \multicolumn{1}{|c|}{\bf BiLSTM} & \multicolumn{1}{|c|}{\bf BiLSTM} & 
\multicolumn{1}{|c|}{\bf LSTM}\\
\multicolumn{1}{|c|}{\bf Stock} & \multicolumn{1}{|c|}{\bf ARIMA} & 
\multicolumn{1}{|c|}{\bf LSTM} & \multicolumn{1}{|c|}{\bf LSTM} & \multicolumn{1}{|c|}{\bf over} & \multicolumn{1}{|c|}{\bf over} & \multicolumn{1}{|c|}{\bf over} \\
& \multicolumn{1}{|c|}{\bf \cite{S. Namini-2018}} & 
\multicolumn{1}{|c|}{\bf } & \multicolumn{1}{|c|}{\bf } & \multicolumn{1}{|c|}{\bf LSTM} & \multicolumn{1}{|c|}{\bf ARIMA} & \multicolumn{1}{|c|}{\bf ARIMA}\\
\hline
N225monthly & 766.45 & 102.49 & 23.13 & -77.43 & -96.98 & -86.66  \\
IXIC.daily & 34.61 & 2.01 & 1.75 & -12.93 & -94.94 & -94.19 \\
IXIC.weekly & 72.53 & 7.95 & 11.53 &  45.03 & -84.10 & -89.03 \\
IXIC.monthly & 135.60 & 27.05 & 8.49 & -68.61 & -93.37 & -80.00  \\
HSI.monthly & 1,306.95 & 172.58 & 121.71 & -29.47 & -90.68 & -86.79 \\
GSPC.daily & 14.83 & 1.74 & 0.62 & -64.36 & -95.81 & -88.26 \\
GSPC.monthly & 55.30 & 5.74 & 4.63 & -19.33 & -91.62 & -89.62 \\
DJI.daily & 139.85 & 14.11 & 3.16 & -77.60 & -97.77 & -89.91 \\
DJI.weekly & 287.60 & 26.61 & 23.05 & -13.37 & -91.98 & -90.74 \\
DJI.monthly & 516.97 & 69.53 & 23.69 & -65.59 & -95.41 & -86.50 \\
IBM.daily & 1.70 & 0.22 & 0.15 & -31.18 & -91.11 & -87.05 \\
\hline
\multicolumn{1}{|c|}{\it Average} & 302.96 & 39.09 & 20.17 & -37.78 & -93.11 & -88.07 \\
\hline
\end{tabular}
\end{center}
\vspace*{-0.3in}
\end{table}

\section{Discussion}
\label{sec:discussion}

As the results show, BiLSTM models outperforms the regular unidirectional LSTMs. It seems that BiLSTMs are able to capture the underlying context better by traversing inputs data twice (from left to right and then from right to left). The better performance of BiLSTM compared to the regular unidirectional LSTM is understandable for certain types of data such as text parsing and prediction of next words in the input sentence. However, it was not clear whether training numerical time series data twice and learning from the future as well as past would help in better forecasting of time series, since there might not exist some contexts, as observable in text parsing. Our results show that BiLSTMs perform better compared to regular LSTMs even in the context of forecasting financial time series data. In order to understand the differences between LSTM and BiLSTM in further details, there are several interesting questions that we can be posed and thus empirically address them, and then learn more about the behavior of these variations of recurrent neural networks and how they work.

\subsection{Loss vs. Batch Steps ($Epoch = 1$)}
In order to compare the loss values for both LSTM and BiLSTM models, we ran the developed scripts on our data and captured the loss values when the learning model fetches the next batch of data. Figure \ref{fig:IBM-loss}(a)-(b) illustrate the plots for the IBM sample data when $Epoch$ is set to $1$, where the y-axis and x-axis represent the loss value and batch steps, respectively. 

As illustrated in Figure \ref{fig:IBM-loss}(a), the loss value starts at $0.061$ and then decreases after fetching the third batch of data for the unidirectional LSTM where the loss values achieves $0.0256$. It implies that after three rounds of fetching the batches of time series data, the loss value remains stable until all batches of data are fetched, where it reaches the loss value of $0.0244$ at its last iteration (i.e., the $42$-{\it th} iteration). 

On the other hand, as shown in \ref{fig:IBM-loss}(b), the loss value starts at $0.0404$, and then, interestingly, its value increases to the highest value (i.e., $0.0874$) on the third round of fetching batches of data. It then starts to decrease slowly after all the batches of data are captured and the parameters are trained. However, unlike the unidirectional LSTM for $Epoch=1$, the BiLSTM model after fetching and learning all the batches of data never reaches the loss value of the counterpart LSTM model (i.e., $0.0256$). This observation may indicate that the BiLSTM model needs fetching more training data to reach the equilibrium in comparison to its unidirectional version (i.e., LSTM). 

As reported in Table \ref{tab:losstable}, the standard deviation calculated for the loss values achieved for both unidirectional LSTM and BiLSTM models when $Epoch = 1$ is $0.007$ and $0.012$, respectively. This indicates that the unidirectional LSTM model reaches the equilibrium faster compared to its counterpart, BiLSTM. The primary reason seems to be directly related to the training the underlying time series processes (first from left-to-right and then right-to-left). As a result, the BiLSTM-based learning model needs to fetch additional data batches to tune its parameters.

\begin{table}
\begin{center}
\caption{Descriptive statistics of loss values for the LSTM and BiLSTM-based models (the IBM stock Data).}
\label{tab:losstable}
\setlength{\tabcolsep}{3pt}
\begin{tabular}{|l|c|c|c|c|c|}
\hline
\multicolumn{1}{|c|}{\bf Model} & \multicolumn{1}{|c|}{\bf Min} & \multicolumn{1}{|c|}{\bf Max} & \multicolumn{1}{|c|}{\bf SD}  & \multicolumn{1}{|c|}{\bf \#Batches} \\ 
\hline
LSTM $Epoch = 1$ & 0.014 & 0.061 & 0.007 & 42 \\
BiLSTM $Epoch = 1$ & 0.026 & 0.087 & 0.012 & 71\\
\hline
LSTM $Epoch = 2 (Round\ 1)$ & 0.013 & 0.048 & 0.005& 41\\
BiLSTM $Epoch = 2 (Round\ 1)$ & 0.025 & 0.184 & 0.02 & 75 \\
\hline
LSTM $Epoch = 2 (Round\ 2)$ & 0.01 & 0.23& 0.004 & 42\\
BiLSTM $Epoch = 2 (Round\ 2)$ & 0.022 & 0.135 & 0.013 & 73 \\
\hline
\end{tabular}
\end{center}
\vspace*{-0.3in}
\end{table}

\subsection{Loss vs. Batch Steps ($Epoch = 2$)}
\begin{figure*}
    \centering
    \subfigure[IBM (LSTM) One Epoch - Loss vs. Batching Step]
    {\includegraphics[scale=0.26]{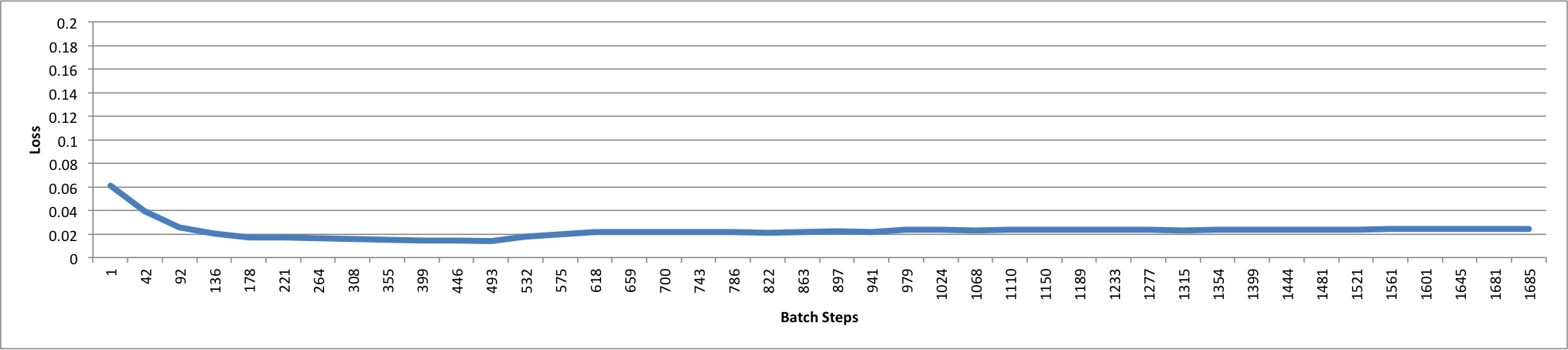}}
       \subfigure[IBM (BiLSTM) One Epoch - Loss vs. Batching Step]
    {\includegraphics[scale=0.26]{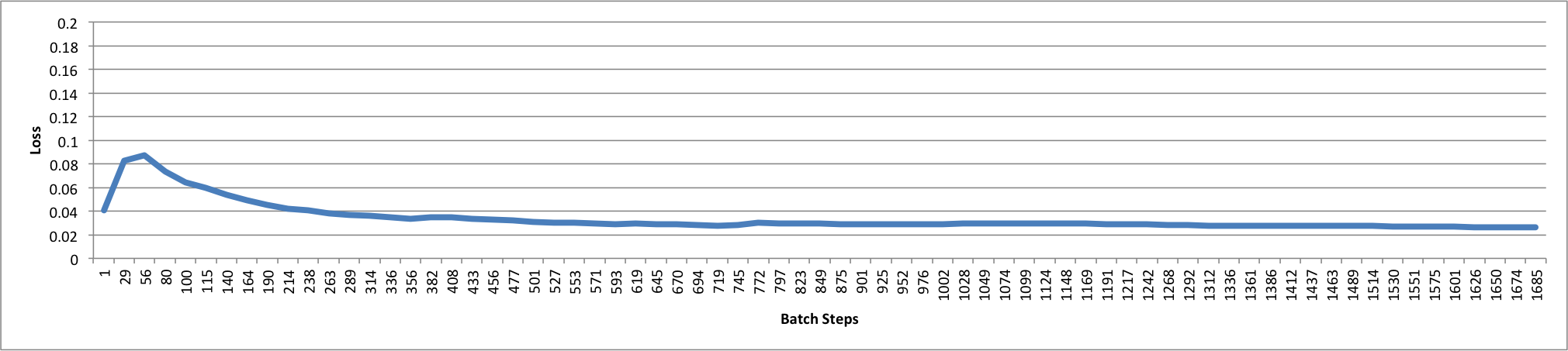}} \\
    \subfigure[IBM (LSTM) Two Epochs (Round 1) - Loss vs. Batching Step]
    {\includegraphics[scale=0.26]{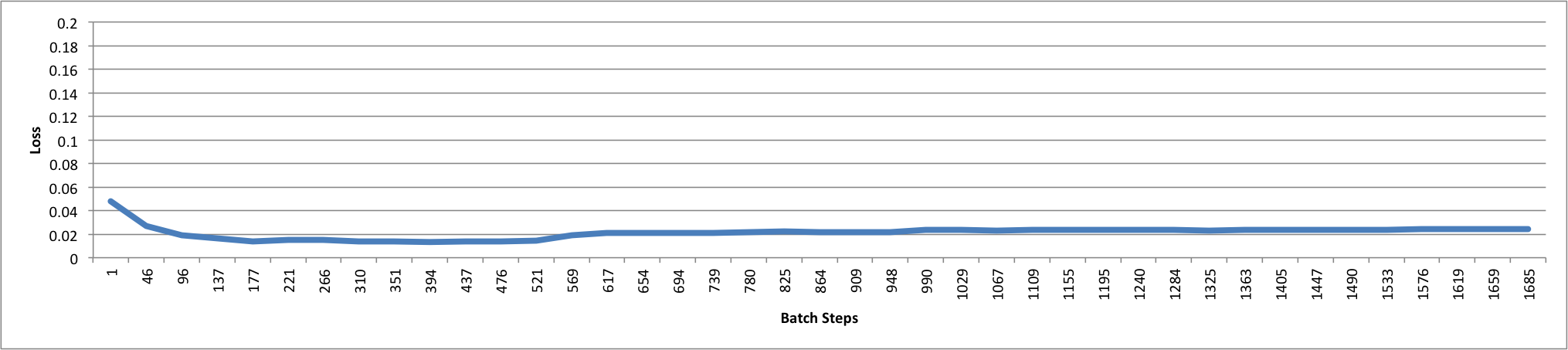}}
        \subfigure[IBM (BiLSTM) Two Epochs (Round 1) - Loss vs. Batching Step]
    {\includegraphics[scale=0.26]{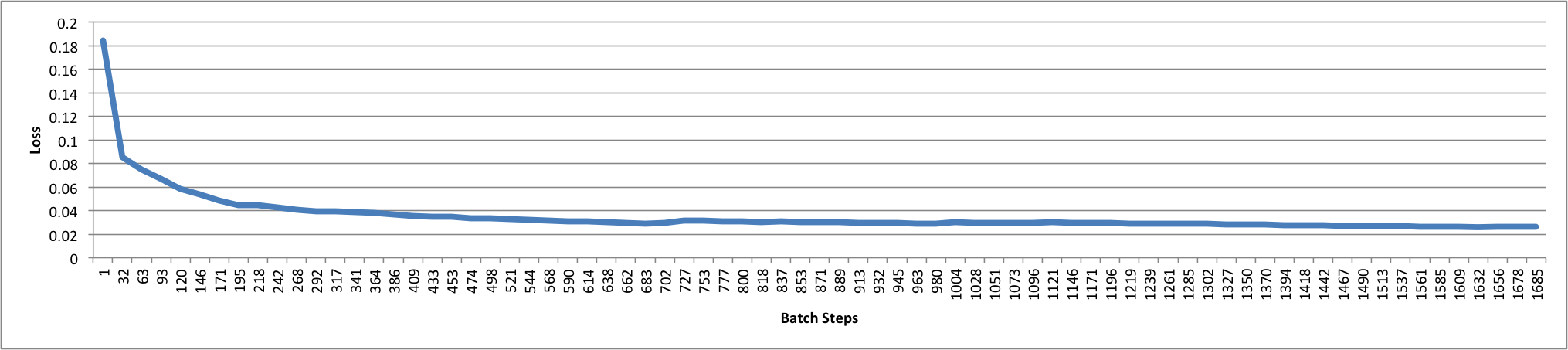}}\\
     \subfigure[IBM (LSTM) Two Epochs (Round 2) - Loss vs. Batching Step]
    {\includegraphics[scale=0.26]{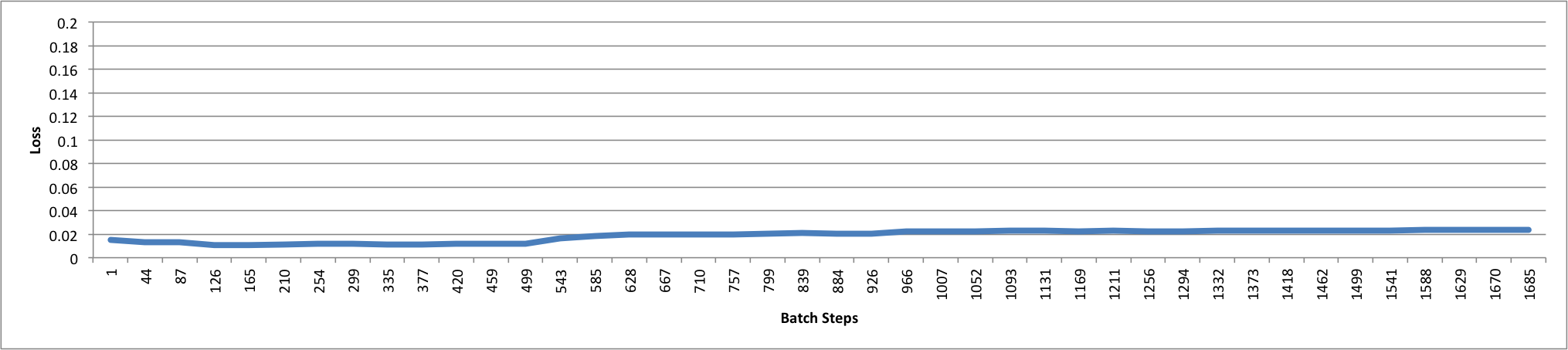}}
         \subfigure[IBM (BiLSTM) Two Epochs (Round 2) - Loss vs. Batching Step]
    {\includegraphics[scale=0.26]{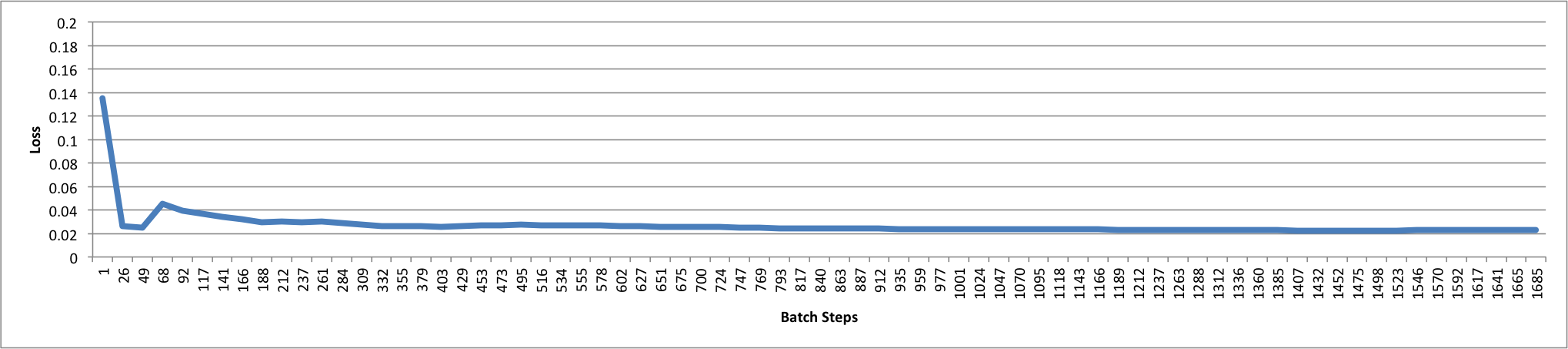}}\\
    \caption{Loss vs. Batch steps for LSTM and BiLSTM - IBM stock forecasting.}
    \label{fig:IBM-loss}
   \vspace*{-0.25in}
    \end{figure*}

The authors also compared the behavioral training of both BiLSTM and LSTM when $Epoch = 2$. Figures \ref{fig:IBM-loss}(c)-(f) illustrate the changes observed in the loss values after fetching each batches of data for the IBM data. 

First the first rounds of Epoch for both BiLSTM and LSTM are compared. Figures \ref{fig:IBM-loss}(c)-(d) illustrate the changes in the loss values for round 1 of training when $Epoch = 2$. We observe similar trends that we obtained for $Epoch = 1$ (Figures \ref{fig:IBM-loss}(a)-(b)) for both LSTM and BiLSTM models for round 1 of $Epoch =2$. More specifically, in round 1 of $Epoch = 2$ of the LSTM model, the loss value starts with $0.048$ and after fetching the 3rd batches is starts to be stabilized, where the loss value is $0.019$. Whereas, for BiLSTM, the loss value starts with $0.184$ and then the loss values start to stabilize after fetching the 8-{\it th} batches for BiLSTM, where the loss value is $0.044$. 

Table \ref{tab:losstable} list the descriptive statistics for both LSTM and BiLSTM for round 1 of $Epoch =2$. As the table reports, a trend similar to  $Epoch =1$ is observed for round 1 of $Epoch =2$, a large standard deviation in the calculated loss values is an indication that BiLSTM requires more data to optimally tune the parameters.

The most intriguing observation is about the changes of loss values in round 2 of $Epoch =2$. Figures \ref{fig:IBM-loss}(e)-(f) demonstrate the trends of changes for loss for both LSTM and BiLSTM. For LSTM, the loss demonstrates a stabilized trend. The loss value starts with $0.015$, remains stable, and after fetching all the batches of data stays at $0.0237$, which shows an insignificant increase. A closer look at Figure \ref{fig:IBM-loss}(c) with \ref{fig:IBM-loss}(e) indicates that the LSTM model has already reached the equilibrium during the first round and during the second round of Epoch nothing valuable is learned.  

On the other hand, the 
trend for round 2 of BiLSTM for $Epoch =2$ does not exhibit a trend similar as observed for the LSTM. As Figure \ref{fig:IBM-loss}(f) illustrates, the training model still keeps continue learning from the data and tuning the parameters. The loss value starts at $0.135$ and then quickly falls into $0.026$ and after a minor fluctuation, it becomes stabilized after fetching the 9-{\it th} batch, where the loss value reaches $0.0295$. A comparison of Figures \ref{fig:IBM-loss}(d) and \ref{fig:IBM-loss}(f) indicates that the BiLSTM model keeps training its parameters after the second round; whereas, the LSTM model stops learning and tuning parameters after the first round. The numerical values and descriptive statistics are reported in Table \ref{tab:losstable}, where the standard deviations for LSTM and BiLSTM are reported as $0.004$ and $0.013$, respectively. Indicating the training needed for optimizing the BiLSTM model in comparison to LSTM.

\subsection{Batch Sizes}
The last column of Table \ref{tab:losstable} reports an interesting phenomenon through which the number of batches, which are considered for the same data by each training model, is reported. According to the experimental data, the LSTM model divided the data into 41 - 42 batches (larger chunks); whereas, the BiLSTM model divided the same data into 71 - 75 batches (smaller chunks). A rational to explain this behavior is the limitation associated with LSTM models in general. Even though these models are capable of ``{\it remembering}'' sequences of data, LSTM-based models have limitations in remembering long sequences. Through a regular LSTM model, since the input data are traversed only once from left to right, a certain number of input items can be fed into the training model. On the other hand, in an BiLSTM model, the training network needs to train not only the input data from left to right, but also from right to left. As a result, the length of training data that can be handled through each batch is almost half of the amount of data learned through each batch by regular LSTM.

\section{Conclusion and Future Work}
\label{sec:conclusion}

This paper reported the results of an experiment, through which the performance and accuracy as well as behavioral training of ARIMA, unidirectional LSTM (LSTM), and bidirectional LSTM (BiLSTM) models were analyzed and compared. The research question targeted by the experiment was primarily focusing on whether training of data from an opposite direction (i.e., right to left), in additional to regular form of training of data (i.e., left to right) had any positive and significant impact on improving the precision of time series forecasting. The results showed that the use of additional layer of training would help in improving the accuracy of forecast by $37.78\%$ percent on average and thus it is beneficial for modeling. We also observed an interesting phenomenon when conducting the behavioral analysis of unidirectional LSTM and BiLSTM models. We noticed that training based on BiLSTM is slower and it takes fetching additional batches of data to reach the equilibrium. This observation indicates that there are some additional features associated with data that might be captured by BiLSTM but unidirectional LSTM models are not capable of exposing them, since the training is only one way (i.e., from left to right). As a result, this paper recommends using BiLSTM instead of LSTM for forecasting problem in time series analysis. This research can be further expanded to forecasting problems for multivariate and seasonal time series.    

\section*{Acknowledgment}
This work is supported in part by National Science Foundation (NSF) under the grants 1821560 and 1723765.


\begin{thebibliography}{00}

\bibitem{Adebiyi-2014} A. A. Adebiyi, A. O. Adewumi, C. K. Ayo, ``Stock Price Prediction Using the ARIMA Model,'' in \emph{UKSim-AMSS 16th International Conference on Computer Modeling and Simulation.}, 2014.

\bibitem{Alonso-2012} A. M. Alonso, C. Garcia-Martos, ``Time Series Analysis - Forecasting with ARIMA models,'' \emph{Universidad Carlos III de Madrid}, Universidad Politecnica de Madrid. 2012.



\bibitem{baldi1999exploiting} P. Baldi, S. Brunak, P. Frasconi, G. Soda, and G. Pollastri, ``Exploiting the past and the future in protein secondary structure prediction,'' \emph{Bioinformatics}, 15(11), 1999.

\bibitem{Brownlee-2017} J. Brownlee, ``How to Create an ARIMA Model for Time Series Forecasting with Python,'' 
2017.

\bibitem{Brownlee-2016} J. Brownlee, ``Time Series Prediction with LSTM Recurrent Neural Networks in Python with Keras,'' 
2016.

\bibitem{Box-1970} G. Box, G. Jenkins, \emph{Time Series Analysis: Forecasting and Control}, San Francisco: Holden-Day, 1970.

\bibitem{Cui2018} Z. Cui, R. Ke, Y. Wang, ``Deep Stacked Bidirectional and Unidirectional LSTM Recurrent Neural Network for Network-wide Traffic Speed Prediction,'' \emph{arXiv:1801.02143}, 2018.

\bibitem{Earnest-2005} A. Earnest, M. I. Chen, D. Ng, L. Y. Sin,  ``Using Autoregressive Integrated Moving Average (ARIMA) Models to Predict and Monitor the Number of Beds Occupied During a SARS Outbreak in a Tertiary Hospital in Singapore,'' in \emph{BMC Health Service Research}, 5(36), 2005.

\bibitem{Fischera-2017} T. Fischera, C. Kraussb, ``Deep Learning with Long Short-term Memory Networks for Financial Market Predictions,'' in \emph{FAU Discussion Papers in Economics 11}, 2017. 

\bibitem{gao2004expanded}, J. Gao, H. Liu, and E.T. Kool, ``Expanded-size bases in naturally sized DNA: Evaluation of steric effects in Watson- Crick pairing,'' \emph{Journal of the American Chemical Society}, 126(38), pp. 11826--11831, 2004. 


\bibitem{Gers-2000} F. A. Gers, J. Schmidhuber, F. Cummins,  ``Learning to Forget: Continual Prediction with LSTM,'' in \emph{Neural Computation} 12(10): 2451-2471, 2000.

\bibitem{Hochreiter-1997} S. Hochreiter, J. Schmidhuber, ``Long Short-Term Memory,'' \emph{Neural Computation} 9(8):1735-1780, 1997.


\bibitem{Huck-2009} N. Huck, ``Pairs Selection and Outranking: An Application to the S\&P 100 Index,'' in \emph{European Journal of Operational Research} 196(2): 819-825, 2009.

\bibitem{Hyndman-2014A} R. J. Hyndman, G. Athanasopoulos, \emph{Forecasting: Principles and Practice}. OTexts, 2014.

\bibitem{Hyndman-2014B} R. J. Hyndman, ``Variations on Rolling Forecasts,'' 
2014.

\bibitem{Kane-2014} M. J. Kane, N. Price, M. Scotch, P. Rabinowitz, ``Comparison of ARIMA and Random Forest Time Series Models for Prediction of Avian Influenza H5N1 Outbreaks,'' \emph{BMC Bioinformatics}, 15(1), 2014. 

\bibitem{Khashei-2011} M. Khashei, M. Bijari, ``A Novel Hybridization of Artificial Neural Networks and ARIMA Models for Time Series forecasting,'' in \emph{Applied Soft Computing} 11(2): 2664-2675, 2011. 


\bibitem{Kim2019} J. Kim, N. Moon, "BiLSTM model based on multivariate time series data in multiple field for forecasting trading area." \emph{Journal of Ambient Intelligence and Humanized Computing}, pp. 1-10.

\bibitem{Krauss-2016} C. Krauss, X. A. Do, N. Huck, ``Deep neural networks, gradient-boosted trees, random forests: Statistical arbitrage on the S\&P 500,'' FAU Discussion Papers in \emph{Economics 03/2016}, Friedrich-Alexander University Erlangen-Nuremberg, Institute for Economics, 2016.

\bibitem{S. Namini-2018} S. S. Namini, N. Tavakoli, and A. S. Namin. "A Comparison of ARIMA and LSTM in Forecasting Time Series." \emph{17th IEEE International Conference on Machine Learning and Applications (ICMLA)}, pp. 1394-1401. IEEE, 2018.


\bibitem{Lee-2017} S. I. Lee, S. J. Seong Joon Yoo, ``A Deep Efficient Frontier Method for Optimal Investments,'' Department of Computer Engineering, Sejong University, Seoul, 05006, Republic of Korea, 2017.

\bibitem{Pang2015} Y. Pang, X. Xue, A.S. Namin, ``Predicting Vulnerable Software Components through N-Gram Analysis and Statistical Feature Selection,'' \emph{International Conference on Machine Learning and Applications {ICMLA}}, pp. 543-548, 2015. 

\bibitem{Patterson-2017} J. Patterson, \emph{Deep Learning: A Practitioner's Approach,} O’Reilly Media, 2017. 


\bibitem{Schmidhuber-2015} J. Schmidhuber, ``Deep learning in neural networks: An overview,'' in \emph{Neural Networks}, 61: 85-117, 2015.

\bibitem{Schuster-1997} M. Schuster, K. K. Paliwal, ``Bidirectional recurrent neural networks'', \emph{IEEE Transactions on Signal Processing}, 45 (11), pp. 2673--2681, 1997.

\bibitem{Tavakoli-2019} N. Tavakoli, ``Modeling Genome Data Using Bidirectional LSTM'' \emph{IEEE 43rd Annual Computer Software and Applications Conference (COMPSAC)}, vol. 2, pp. 183-188, 2019.


\bibitem{Tavakoli-IO-2019} N. Tavakoli, D. Dong, and Y. Chen, "Client-side straggler-aware I/O scheduler for object-based parallel file systems." \emph{Parallel Computing}, pp. 3-18,82, 2019.










 
 
 
























\end{thebibliography}
\end{document}